\documentclass[letterpaper, 10 pt, conference]{ieeeconf}

\usepackage{graphicx}
\usepackage{xcolor}
\definecolor{mydarkblue}{rgb}{0,0.08,0.45}
\usepackage[colorlinks = true,
            linkcolor = mydarkblue,
            urlcolor  = mydarkblue,
            citecolor = mydarkblue,
            anchorcolor = mydarkblue]{hyperref}
\usepackage{amsmath}
\usepackage{amssymb}
\usepackage{mathtools}
\usepackage{bbm}
\usepackage{glossaries}
\usepackage{booktabs}
\usepackage{caption}
\usepackage{subcaption}
\usepackage{dsfont}
\usepackage{eso-pic}

\newcommand\copyrightnotice{%
  \AddToShipoutPictureBG*{%
    \AtPageLowerLeft{%
      \put(\LenToUnit{0.5\paperwidth},\LenToUnit{0.8cm}){%
        \makebox[0pt][c]{%
          \parbox[b]{\textwidth}{\scriptsize\centering
            \copyright~2026 IEEE. Personal use of this material is permitted.
            Permission from IEEE must be obtained for all other uses, in any
            current or future media, including reprinting/republishing this
            material for advertising or promotional purposes, creating new
            collective works, for resale or redistribution to servers or
            lists, or reuse of any copyrighted component of this work in
            other works.}%
        }%
      }%
    }%
  }%
}

\IEEEoverridecommandlockouts
\glsdisablehyper

\newacronym{drl}{DRL}{Deep Reinforcement Learning}
\newacronym{mlp}{MLP}{Multilayer Perceptron}
\newacronym{lstm}{LSTM}{Long Short-Term Memory}
\newacronym{gru}{GRU}{Gated Recurrent Unit}
\newacronym{cnn}{CNN}{Convolutional Neural Network}
\newacronym{pomdp}{POMDP}{Partially Observable Markov Decision Process}
\newacronym{bptt}{BPTT}{Backpropagation Through Time}
\newacronym{hmf}{HMF}{Hadamard Memory Framework}
\newacronym{ppo}{PPO}{Proximal Policy Optimization}
\newacronym{lsmm}{LSMM}{Local Spatial Matrix Memory}
\newacronym{mjx}{MJX}{MuJoCo XLA}
\newacronym{sru}{SRU}{Spatially-Enhanced Recurrent Unit}

\title{\LARGE \bf
Remember Where You Walk:\\Learning Spatial Memory for End-to-End Legged Robot Exploration
}

\author{Leon Magnus$^{*,1}$, Nico Bohlinger$^{*,1}$, Jan Peters$^{1,2,3,4}$, Oleg Arenz$^{1}$%
 \thanks{This project was funded by National Science Centre Poland in the Weave programme UMO-2021/43/I/ST6/02711, by the German Science Foundation (DFG) under grant number PE 2315/17-1 and by the European Union’s Horizon Europe research and innovation
programme (Grant agreement ID: 101298421). This project has been supported
by a hardware donation by NVIDIA through the Academic Grant Program. * Equal contribution. $^{1}$ Technical University of Darmstadt, Germany. $^{2}$ Robotics Institute Germany (RIG). $^{3}$ German Research Center for AI (DFKI).$^{4}$ hessian.AI}
 }%

\begin{document}

\maketitle
\copyrightnotice
\thispagestyle{empty}
\pagestyle{empty}

\begin{abstract}
 Humanoids and other legged robots operating in complex environments must remember information that leaves their field of view and only becomes relevant much later. Reinforcement learning policies address this using memory networks such as LSTMs that are trained end-to-end with the policy to memorize such task-relevant aspects. Yet, such architectures often fail to preserve sufficient information over long horizons. In this work, we propose a network architecture based on Hadamard memory that specifically targets robot navigation. Our \gls{lsmm} enables RL agents to store generic task-relevant information at specific, position-based memory cells, enabling flexible and spatially grounded memorization. We evaluate \gls{lsmm} on the task of autonomous maze exploration, where it matches the maze coverage of strong recurrent and
finite-history baselines while being the only architecture that
remembers the maze layout, reducing the reconstruction error threefold. We provide videos and code for our experiments at: \url{https://leonmgns.github.io/RememberWhereYouWalkWebiste/}

\end{abstract}

\section{Introduction}

Thanks to breakthroughs in \gls{drl}, legged robots can now traverse rough terrain, jump over obstacles, and navigate toward goal positions using policies trained in highly-parallelized simulations or directly in the real world \cite{zhuang2023, cheng2023, hoeller2024anymal, stasica2025bridge, bohlinger2025gait}.
These policies usually follow velocity commands or short-horizon waypoints \cite{bohlinger2024onepolicy, rudin2022advanced}, leaving the higher level planning to a human operator and making the system not fully autonomous.
Autonomous exploration of an unknown environment is a fundamentally different problem.
Now, the robot must not only reason and control its body in a local frame, but also decide where to go and remember where it has already been, to navigate and explore efficiently \cite{yamauchi1997frontier}.
This turns locomotion into a long-horizon partially observable navigation problem in which the current local observation is not enough \cite{mirowski2016learning}.

Memory-augmented policies address this problem by leveraging an observation-action history with architectures such as \glspl{lstm} \cite{hochreiter1997long}, \glspl{gru} \cite{cho2014properties}, Transformers \cite{vaswani2017attention}, or simply concatenating history for an \gls{mlp} \cite{ji2022concurrent}.
However, unstructured recurrent states, through \glspl{lstm} or \glspl{gru}, easily mix or overwrite spatially distinct information, as credit-assignment is tricky, and finite-context approaches, like Transformers or concatenation, discard old observations in less than a second with the typical history lengths \cite{ji2022concurrent, li2026online}.
Classical navigation systems avoid this problem by explicitly registering and using observations in spatial maps, but typically separate mapping, planning, and locomotion into individual components, losing out on potential end-to-end efficiency \cite{miki2022elevation, aditya2025robust, mattamala2025wild}.

\begin{figure}[t]
    \centering
    \includegraphics[width=1.0\linewidth]{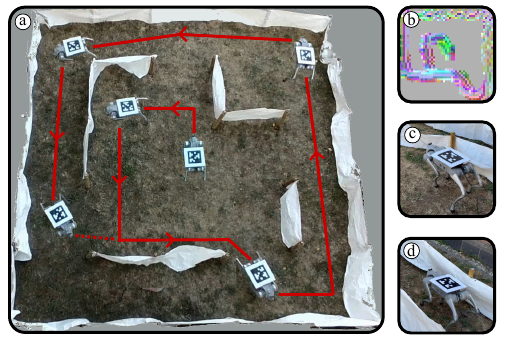}
    \caption{
    (a) The Unitree Go2 autonomously explores a previously unseen $5\times5\,\mathrm{m}$ outdoor maze. The red line shows the walked path.
    (b) The matrix memory at the end of the run.
    The policy has written into the cells along its path, shown with the
    three memory channels interpreted as RGB colors, while unwritten cells keep their initial value in gray.
    (c) and (d) Shows the robot navigating through the maze.
    }
    \label{fig:real_world}
\end{figure}

In this work, we introduce \gls{lsmm} and evaluate it for end-to-end maze exploration.
The layout of our matrix memory is aligned with the environment's frame and indexed using a position estimate.
At every step, the policy modifies only a small window in the memory around the robot's position through learned calibration and update matrices, while the rest of the memory is not updated.
A \gls{cnn} \cite{lecun1989backpropagation} encoder transforms the spatial memory into features used by both the policy and an auxiliary maze-reconstruction head.
This combines the information persistence of index-based updates and the locality of map-based navigation with learned latent representations and end-to-end locomotion control, without any explicit path planning or waypoint commands.

We formulate long-horizon, goal-free maze exploration as an end-to-end locomotion problem.
We train our method for the Unitree Go2 quadruped in procedurally generated mazes and compare \gls{lsmm} against unstructured recurrent and finite-history architectures.
Our experiments provide comprehensive empirical evaluations in simulation and deployment in the real-world, demonstrating \gls{lsmm}'s capabilities on maze reconstruction, exploration efficiency, robustness to localization noise, and generalization to unseen mazes.

\section{Related Work}
\label{sec:related_work}

Partial observability in \gls{drl} is commonly addressed by conditioning the policy on a finite observation history or a recurrent latent state \cite{al2024time}.
Recurrent architectures such as \glspl{lstm} and \glspl{gru} have been widely used for navigation tasks \cite{hochreiter1997long,cho2014properties,mirowski2016learning,yang2025spatially}, while attention-based policies can leverage longer observation sequences \cite{vaswani2017attention,liulocoformer}.
Le et al.\ show that \glspl{lstm}, \glspl{gru}, and many other recurrent architectures update their memory in the same underlying way, calibrating the previous memory element-wise and then writing new information additively \cite{le2025stable}.
They formalize this update rule as the \gls{hmf} and derive Stable Hadamard Memory, whose calibration keeps the memory from decaying or exploding over long horizons.
However, all of these memories aggregate observations over time and maintain no explicit link between the stored information and the spatial location where it was acquired.

Spatial neural memories address this limitation by aligning the latent state with the geometry of the environment.
Neural Map stores features in a two-dimensional memory indexed by the agent's position for maze navigation \cite{parisotto2018neural}.
MapNet similarly constructs a self-centered latent map and learns to register new observations within it \cite{henriques2018mapnet}.
Active Neural SLAM combines a learned mapping module with global and local policies, but keeps an explicit analytical planner between the map and the executed actions \cite{chaplot2020learning}.
Our \gls{lsmm} keeps the spatial indexing of these methods but does not need the analytical planner of Active Neural SLAM, since its memory features enter the locomotion policy directly.
Unlike Neural Map, which writes a single feature vector into the agent's current cell, it retrieves a whole window of cells around the robot's estimated position, applies the \gls{hmf} update rule inside this window, and leaves the rest of the memory unchanged.
The memory and the locomotion policy are trained jointly, and an auxiliary head reconstructs the maze layout from the memory features, which encourages it to retain task-relevant information about the maze.

Classical legged-navigation stacks put the locomotion policy as the last layer in their pipeline, where it executes velocity or waypoint commands produced by separate localization, mapping, and planning modules \cite{miki2022elevation,aditya2025robust,mattamala2025wild}.
Learning-based approaches tried to merge parts of this pipeline.
Rudin et al.\ train a single policy that reaches a commanded target position within a given time \cite{rudin2022advanced}, and Zhang et al.\ learn a navigation policy that steers a separate locomotion policy to such a target \cite{zhang2024resilient}.
In both cases the target position itself still comes from an external planner or operator.
For exploration this planner typically builds an occupancy map and repeatedly picks the next target on the frontier between the mapped and unexplored space of the environment \cite{yamauchi1997frontier}.
We remove this remaining component, since the spatial memory takes over the role of the occupancy map and the decision where to move next is learned jointly with the locomotion.

\section{Spatial Memory for End-to-End\\Legged Robot Exploration}

We propose the \gls{lsmm}, a spatially indexed latent memory that allows a single policy to remember its environment, decide where to go, and control the robot's joints at the same time.
First, we formalize the exploration task and the interface of the policy.
Then, we introduce the local memory update of \gls{lsmm} and describe how memory, encoders, and policy are trained jointly with \gls{ppo} \cite{schulman2017proximal} and an auxiliary maze reconstruction objective.

\subsection{Problem Formulation}
\label{sec:problem}

We consider the autonomous exploration of unseen mazes with the Unitree
Go2 quadruped.
The robot starts at the center of a $10\times10\,\mathrm{m}$ maze built
from a $5\times5$ grid of $2\times2\,\mathrm{m}$ cells and must visit as
many cells as possible within an episode of $70\,\mathrm{s}$.
At deployment, the policy receives no goal position, waypoint, or velocity
command.
Since the current observation does not reveal which parts of the maze the
robot has already visited, the task is a \gls{pomdp} in which the policy
must remember information over time.
We therefore learn a memory-augmented policy
$\pi_\theta(\mathbf{a}_t \mid \mathbf{o}_t, \mathbf{M}_t)$ that maximizes
the expected discounted return
$\mathbb{E}\left[\sum_t \gamma^t r_t\right]$, where $\mathbf{M}_t$ is the
spatial memory introduced in \autoref{sec:lsmm} and $r_t$ is the
reward described in \autoref{sec:training}.

The observation $\mathbf{o}_t$ consists of a three-dimensional waypoint command $\mathbf{c}_t$, which is only used in early training and set to zero otherwise, the proprioceptive state $\mathbf{o}^{\text{prop}}_t$ comprising joint positions, joint velocities, the previous action, the angular velocity of the trunk, and the projected gravity vector, and an exteroceptive observation $\mathbf{o}^{\text{h}}_t$ of 250 terrain height samples around the trunk.
$\hat{\mathbf{p}}_t=(\hat{x}_t,\hat{y}_t,\hat{\psi}_t)$ of the robot's
position and heading in the environment frame.
The critic has access to privileged information, namely the linear
velocity of the trunk, its height above ground, the contact states of the
feet, the ground-truth position, the per-cell visit counts, and a wider map of height samples.
The actions $\mathbf{a}_t$ are target joint position offsets to a nominal
standing configuration, tracked by a PD controller at $50\,\mathrm{Hz}$. 

\subsection{Policy Architecture}
\label{sec:architecture}

\begin{figure}[t]
    \centering
    \includegraphics[width=\linewidth]{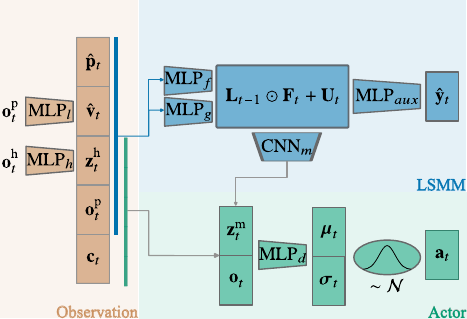}
    \caption{Actor architecture with the proposed \gls{lsmm} memory module. Both branches share $\mathbf{z}_t^{\text{h}}$ and $\mathbf{o}_t^{\text{p}}$, while the memory branch additionally uses noisy position and velocity estimates $\hat{\mathbf{p}}_t$ and $\hat{\mathbf{v}}_t$ to update the matrix memory $\mathbf{L}_t$. A CNN then extracts $\mathbf{z}_t^{\text{m}}$ from $\mathbf{L}_t$ and an auxiliary \gls{mlp} predicts the maze wall layout $\hat{\mathbf{y}}_t$. The actor maps $\mathbf{z}_t^{\text{m}}$ and its observation $\mathbf{o}_t$ to $\boldsymbol{\mu}_t$ and $\boldsymbol{\sigma}_t$, from which $\mathbf{a}_t$ is sampled.}
    \label{fig:lsmm-architecture}
\end{figure}

\autoref{fig:lsmm-architecture} shows the policy architecture, which
consists of two parts, the memory module and the actor.
Since the Go2's IMU cannot measure linear velocity directly, an \gls{mlp}
first estimates it from the proprioceptive observation as
$\hat{\mathbf{v}}_t = l(\mathbf{o}^{\text{p}}_t)$.
A second \gls{mlp} encodes the height samples into a feature vector
$\mathbf{z}^{\text{h}}_t = h(\mathbf{o}^{\text{h}}_t)$.
Then, the memory module takes
$[\mathbf{o}^{\text{p}}_t, \mathbf{z}^{\text{h}}_t,
\hat{\mathbf{v}}_t, \hat{\mathbf{p}}_t]$ and writes them into the
\gls{lsmm}, as described in \autoref{sec:lsmm}.
A \gls{cnn} then encodes the updated memory into a feature vector
$\mathbf{z}^{\text{m}}_t = m(\mathbf{M}_t)$.
Finally, the action is sampled from a Gaussian distribution
$$
\mathbf{a}_t \sim \mathcal{N}\big(\boldsymbol{\mu}_t,
(\boldsymbol{\sigma}_t^2)\big), \quad (\boldsymbol{\mu}_t, \boldsymbol{\sigma}_t) =
d([\mathbf{c}_t, \mathbf{o}^{\text{p}}_t, \mathbf{z}^{\text{h}}_t,
\mathbf{z}^{\text{m}}_t]).
$$
Together, these networks form the policy $\pi_\theta$, where $\theta$
collects the parameters of the velocity estimator $l$, the encoders $h$
and $m$, the memory write functions $f$ and $g$, and the actor $d$.

We train with an asymmetric actor-critic setup \cite{pinto2018asymmetric}.
The critic is an \gls{mlp} without memory.
It receives the privileged observations from \autoref{sec:problem}, where
the per-cell visit counts summarize the exploration progress that the
actor has to remember.
An auxiliary head predicts the wall layout of the maze from the actor
features at every step.
The full layout can only be recovered from information gathered earlier in
the episode, so predicting it requires the memory to store what the robot
has seen.
At deployment, the critic and the wall prediction are dropped, and only
$\pi_\theta$ runs on the robot.

\subsection{Local Spatial Matrix Memory}
\label{sec:lsmm}

The \gls{lsmm} is a tensor $\mathbf{M}_t \in [0,1]^{H \times W \times C}$
whose first two dimensions are aligned with the environment frame.
We use $H \times W = 50 \times 50$ cells with $C = 3$ channels, so one
cell covers $0.2 \times 0.2\,\mathrm{m}$ of the maze.
The channels carry learned features without a predefined meaning, and all
entries are initialized to $0.5$ at the beginning of an episode.

Updating the full memory at every step, as done by the recurrent
architectures in the \gls{hmf} \cite{le2025stable}, would constantly
rewrite information about locations the robot currently cannot observe.
Therefore, \gls{lsmm} only updates a window $\mathcal{W}_t$ of
$5 \times 5$ cells centered on the memory cell of the position estimate
$\hat{\mathbf{p}}_t$, corresponding to $1 \times 1\,\mathrm{m}$ around the
robot.
Let $\mathbf{L}_{t-1} = \mathbf{M}_{t-1}[\mathcal{W}_t]$ denote the
current content of this window.
Two \glspl{mlp} $f$ and $g$ compute a calibration tensor $\mathbf{F}_t$
and an update tensor $\mathbf{U}_t$ as
\begin{equation}
\begin{aligned}
    \mathbf{F}_t &= f\big(\big[\mathbf{o}^{\text{p}}_t,
    \mathbf{z}^{\text{h}}_t, \hat{\mathbf{v}}_t, \hat{\mathbf{p}}_t,
    \mathbf{L}_{t-1}\big]\big), \\
    \mathbf{U}_t &= g\big(\big[\mathbf{o}^{\text{p}}_t,
    \mathbf{z}^{\text{h}}_t, \hat{\mathbf{v}}_t, \hat{\mathbf{p}}_t,
    \mathbf{L}_{t-1}\big]\big),
\end{aligned}
\label{eq:write_functions}
\end{equation}
where both tensors match the shape of the window and are bounded to
$(0,1)$ by a sigmoid.
Through $\mathbf{L}_{t-1}$ the networks read the window before updating it,
so $\mathbf{F}_t$ acts as a forget gate on the stored content and
$\mathbf{U}_t$ writes new information into it.
Following the update rule of the \gls{hmf}, the window content is then
overwritten with
\begin{equation}
    \tilde{\mathbf{L}}_t
    = \mathbf{L}_{t-1} \odot \mathbf{F}_t + \mathbf{U}_t,
    \qquad
    \mathbf{L}_t
    = \frac{\tilde{\mathbf{L}}_t}{\max(\tilde{\mathbf{L}}_t) + \epsilon},
    \label{eq:lsmm}
\end{equation}
where $\odot$ denotes the element-wise product and the normalization with
$\epsilon = 0.02$ keeps all entries in $[0,1]$.
All entries outside $\mathcal{W}_t$ remain unchanged.

Observations made in different parts of the maze therefore modify different
parts in the memory, and a written cell keeps its latent exactly until
the robot returns to the corresponding location, independent of how much
time passes in between.
The calibration through $\mathbf{F}_t$ lets the policy decide how much of
the previously stored information to keep when it revisits a location, and
the spatial layout of $\mathbf{M}_t$ allows a direct visual comparison
with the true maze, which we use in \autoref{sec:results}.

\subsection{RL Training Setup in Simulation}
\label{sec:training}

We train policy and critic with \gls{ppo} in $5\,000$ parallel \gls{mjx} \cite{todorov2012mujoco} environments with $300\,000$ pre-generated
maze layouts.
The full objective combines the clipped surrogate and value losses of
\gls{ppo} with the maze reconstruction and velocity estimation losses
\begin{equation}
\begin{aligned}
    \mathcal{L}
    &= \mathcal{L}_{\text{PPO}}
    + \lambda_{\text{maze}} \mathcal{L}_{\text{maze}}
    + \lambda_{\text{vel}} \mathcal{L}_{\text{vel}}, \\
    \mathcal{L}_{\text{maze}} &= -\frac{1}{n_{\text{w}}}
    \sum_{i=1}^{n_{\text{w}}} \big[\, y_{t,i} \log \hat{y}_{t,i}
    + (1-y_{t,i}) \log(1-\hat{y}_{t,i}) \,\big], \\
    \mathcal{L}_{\text{vel}} &=
    \lVert \mathbf{v}_t - \hat{\mathbf{v}}_t \rVert^2,
\end{aligned}
\label{eq:loss}
\end{equation}
where $\mathbf{y}_t \in \{0,1\}^{n_{\text{w}}}$ indicates which of the
$n_{\text{w}}$ possible wall segments of the maze exist,
$\hat{\mathbf{y}}_t$ contains the wall probabilities predicted by the
auxiliary head, and $\mathbf{v}_t$ is the simulated linear velocity of
the trunk.
The gradients of $\mathcal{L}_{\text{maze}}$ reach the write functions $f$
and $g$ through the memory encoder $m$ and train the memory to store the
maze layout.
The exploration task demands long-horizon credit assignment, as reaching
the last unexplored cells at the end of an episode depends on decisions
made many seconds earlier.
Therefore, we use a high discount factor of $\gamma = 0.995$.
\autoref{tab:hyperparameters} lists all learning hyperparameters.

\subsection{Rewards}
Each training episode is either a goal episode, in which the robot is
commanded toward a neighboring maze cell, or an exploration episode, in
which it receives no command and must cover as much of the maze as possible.
The reward $r_t$ consists of task rewards, which depend on this episode
type, and shaping penalties for the gait and the navigation.
In exploration episodes, the robot collects a reward for being in rarely
visited cells, which decays with the time already spent in the current
cell, and a reward for the overall maze coverage.
In goal episodes, it is instead rewarded for moving toward the commanded
cell and reaching it.
The shaping penalties punish stalling, walking close to walls, joint
limits, torques, action rates, and asymmetric gaits.
\autoref{tab:rewards} lists all terms and coefficients.
\begin{figure}[!t]
    \centering
    \includegraphics[]{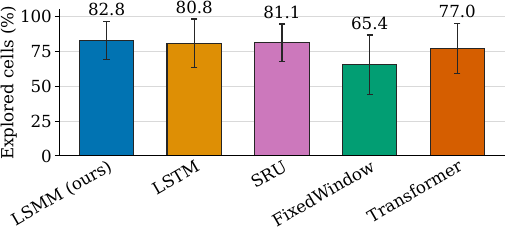}
    \caption{
    Percentage of explored cells at the end of an episode.
    Bars show the mean over 100 unseen evaluation mazes and three seeds per
    architecture, whiskers the standard deviation.
    All architectures reach a comparable coverage.
    }
    \label{fig:exploration}
\end{figure}
\subsection{Curricula}
Learning to walk and to explore at the same time is difficult, as the
exploration rewards provide no useful signal while the robot cannot walk
yet.
Therefore, we bootstrap training with goal episodes, in which the command
$\mathbf{c}_t$ contains the position of a reachable neighboring maze cell
in the robot's base frame.
The least visited neighboring cell is selected to already bias the policy
toward exploration, and once it is reached, the next goal is sampled
immediately.
The last entry of $\mathbf{c}_t$ flags the episode type, and in
exploration episodes the command is the constant vector $[0,0,1]$, so the
policy must decide on its own where to move.
The fraction of exploration episodes grows with a performance-based
curriculum coefficient \cite{ai2025towards,bohlinger2025multi}, which increases when the robot covers at least
$80\,\%$ of the maze in a goal episode or $60\,\%$ in an exploration
episode, and decreases on low coverage or early termination.
The same coefficient ramps up the shaping penalties, so that they do not
suppress the learning of exploration early in training.
A second coefficient grows the same way and controls the fraction of the
maze's walls that are present, so the policy first learns to walk and
follow goals in open spaces before it faces more and more complicated mazes.
At the end of training, effectively all episodes are exploration episodes,
and the deployed policy never depends on an external goal.

\subsection{Domain Randomization}
To enable sim-to-real transfer to the real robot, we apply extensive domain
randomization during training.
We randomize the robot model (trunk mass, center of mass, feet sizes,
collision geometry, joint damping, armature, and friction), the
environment (gravity, ground friction, and contact compliance),
the actuation (motor strength and action delays), and the initial state of
the robot.
We further add noise to all proprioceptive observations, distort the
height samples with noise, tilts, and delays of up to $60\,\mathrm{ms}$,
perturb the position estimate by up to $\pm0.20\,\mathrm{m}$,
corresponding to one memory cell, and push the robot with velocities of up to $0.3\,\mathrm{m/s}$.

\section{Experimental Results}
\label{sec:results}

Our experiments evaluate how \gls{lsmm} explores compared to unstructured memory architectures, what the different memories actually retain about the environment, and whether the learned behavior transfers to the real robot.

We compare \gls{lsmm} against an \gls{lstm}, a \gls{sru} \cite{yang2025spatially}, a Transformer, and an
\gls{mlp} that receives a fixed window of stacked past observations.
All baselines receive the same observations, including the position
estimate, are trained with the same auxiliary losses and the
hyperparameters of \autoref{tab:hyperparameters}, and differ from
\gls{lsmm} only in their memory mechanism.
To keep the comparison fair, we give every approach the same hardware
budget: for each architecture, we choose the largest memory capacity
for which training still fits on a single NVIDIA RTX~3090 with
24\,GB of VRAM.
This results in a hidden state of size $1\,000$ for the \gls{lstm}, \gls{sru} and
a context of the 10 most recent observations for the Transformer and
the fixed-window \gls{mlp}. Larger configurations exceed the available
GPU memory.
We train three seeds per architecture and evaluate each deterministically
in 100 environments on previously unseen mazes.

\begin{figure}[t]
    \centering
    \includegraphics[width=0.95\linewidth]{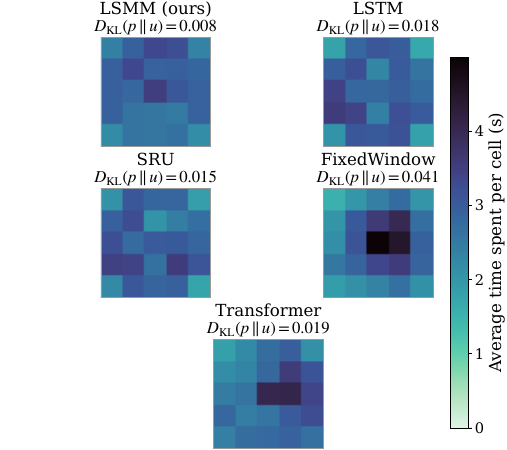}
    \caption{
    Average number of visits per maze cell across all evaluation episodes.
$D_{\text{KL}}(p \,\|\, u)$ measures the divergence between the
normalized visit distribution $p$ and a uniform distribution $u$, where
lower values mean more even exploration.
\gls{lsmm} spreads its visits most evenly over mazes.
    }
    \label{fig:heat_map}
\end{figure}
\subsection{Simulation}

\autoref{fig:exploration} shows the fraction of explored cells at the end of an episode.
All architectures reach a comparable coverage of $65.4$ to $82.8\,\%$, so
pure exploration performance in our mazes does not require a spatial
memory.
The visit distributions in \autoref{fig:heat_map} reveal a first
difference, as \gls{lsmm} spreads its visits most evenly over the maze,
with a KL divergence to the uniform distribution of $0.008$ compared to $0.015$ to 
$0.018$ to $0.019$ for the \gls{sru}, \gls{lstm} and Transformer, while the fixed window architecture concentrates on the cells around the start.

\begin{figure}[t]
    \centering
    \includegraphics[]{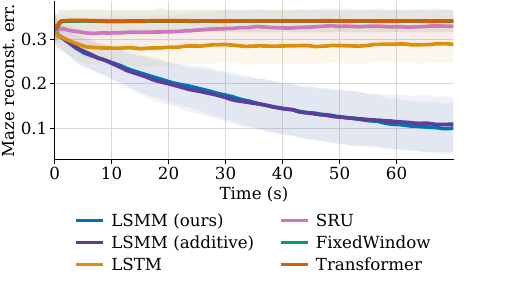}
    \caption{
    Maze reconstruction error over the course of an episode.
    Lines show the mean over 100 unseen evaluation mazes and three seeds, shaded areas the standard deviation.
    \gls{lsmm} is the only architecture that significantly reduces the error while exploring.
    }
    \label{fig:bce_error}
\end{figure}

\begin{figure}[t]
    \centering
    \includegraphics[]{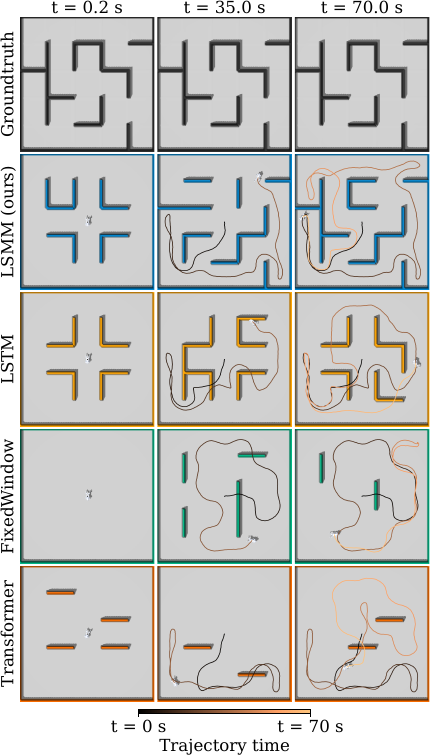}
    \caption{
    Predicted maze layout at three times during an evaluation episode.
    The trajectory walked up to each time is overlaid, colored from black to orange by trajectory time.
    The prediction of \gls{lsmm} fills in wall by wall as the robot explores and closely matches the ground truth at the end of the episode, while the baselines struggle to remember the walls in the areas they have been before.
    }
    \label{fig:maze_predictions}
\end{figure}

However, coverage alone does not show what a policy has remembered.
\autoref{fig:bce_error} tracks the maze reconstruction error over the
episode, and \gls{lsmm} is the only architecture that significantly reduces it while exploring, from $0.33$ to $0.1$, more than three times lower than any baseline.
\autoref{fig:maze_predictions} visualizes this difference, where the
maze predicted by \gls{lsmm} fills in wall by wall as the robot walks,
until it closely matches the ground truth, while the baselines either never predict more than a few walls near their current position or hallucinate walls in free spaces.
When visually inspecting the behaviour of the policies, the baselines explore more reactively and forget the areas they have seen before, which can even lead to circling around walls in loops.
\gls{lsmm} avoids returning to previously explored areas by building a persistent map of the maze.
It rarely hallucinates walls and keeps true walls in memory as long as it has explored the cell enough.
The policies act mostly restless and have a tendency to pace from cell to cell at $0.8-1.2\,\mathrm{m/s}$ to maximize the intrinsic visitation reward.


\begin{figure}[t]
    \centering
    \includegraphics[]{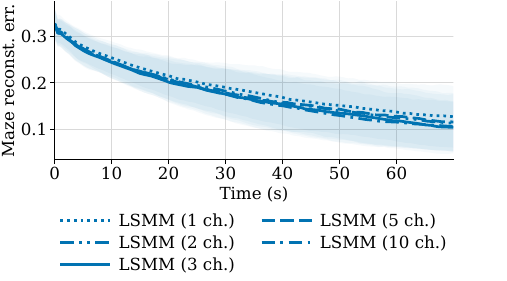}
    \caption{
Maze reconstruction error of \gls{lsmm} for memory sizes of $50 \times 50 \times C$ with $C \in \{1, 2, 3, 5, 10\}$ channels. Lines show the mean over 100 unseen evaluation mazes and three seeds, shaded
areas the standard deviation. All memory sizes perform comparably.
}
    \label{fig:channel_abl}
\end{figure}

\begin{figure}[t]
    \centering
    \includegraphics[]{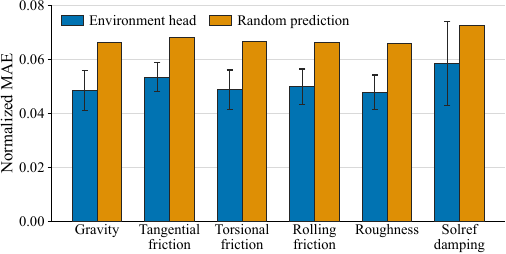}
    \caption{
    Auxiliary prediction of spatially varying environment properties.
    Bars show the normalized mean absolute error over all visited cells at the final time step.
    }
    \label{fig:env_head}
\end{figure}

To test whether the spatial memory can also support predictions beyond the maze layout, we train an auxiliary head, similar to the maze prediction head, to estimate locally varying environment properties, including gravity, ground friction, roughness, and ground damping.
For each cell in a maze, we sample each property uniformly from a range scaled by the performance-based curriculum.
\autoref{fig:env_head} shows consistently lower error than random guessing.
Therefore, the policy can not only store where it has been but also aspects of the world it experienced there.
We suspect that with a larger latent size for the memory cells, better estimations and even more predicted properties would be possible.

\subsection{LSMM Memory Ablation}
We ablate two design choices of the \gls{lsmm} memory.
First, \gls{hmf} rescales the already written memory with a calibration tensor $\mathbf{F}_t$ at every write step.
To evaluate its relevance, \autoref{fig:bce_error} compares \gls{lsmm} with an additive variant that updates the memory only with the update tensor $\mathbf{G}_t$.
Both variants reduce the reconstruction error similarly, which indicates that calibration is not required for remembering the maze layout in our setting. A likely reason is that the mazes are static during an episode, so once a wall
is written to memory it never needs to be revised, and purely additive updates suffice.

Second, \autoref{fig:channel_abl} shows the reconstruction error for different
numbers of memory channels.
All configurations lie within each other's standard deviation, and the mean
error of a single channel differs only slightly from that of $3$ and $10$
channels without being significant.
The reconstruction performance is thus largely insensitive to the memory size,
suggesting that a single channel per cell already provides enough capacity to
encode the wall layout.

\subsection{Real World}
\begin{figure}[t]
    \centering
    \includegraphics[width=0.7\linewidth]{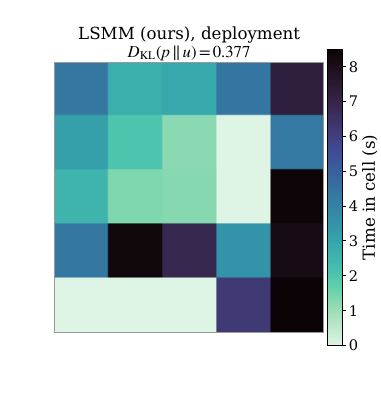}
    \caption{
    Time spent in each cell by our deployed \gls{lsmm} policy. Compared to the simulation
    results, this deployment run distributes its visits less uniformly and follows a more
    repetitive pattern, which is also visible in the matrix memory in
    \autoref{fig:real_world}~(b).
}
    \label{fig:heat_map_real}
\end{figure}
\begin{figure}[t]
    \centering
    \includegraphics[]{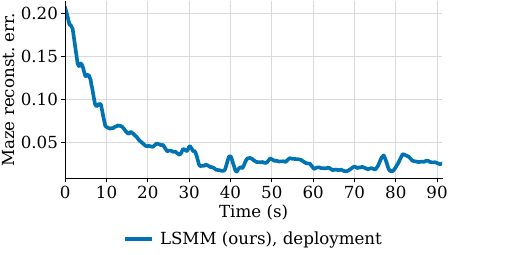}
    \caption{
    Maze reconstruction error over deployment time of our \gls{lsmm} policy. The line shows the reconstruction error of the deployment run shown in \autoref{fig:real_world}. The policy reduces the error significantly within the first $30\,\mathrm{s}$. }
    \label{fig:bce_real}
\end{figure}

We deploy the \gls{lsmm} policy on the Unitree Go2 in an unseen outdoor maze, as shown in \autoref{fig:real_world}.
We reduce the size of the maze from $10\times10\,\mathrm{m}$ to $5\times5\,\mathrm{m}$ due to constraints on the available space. We run the policy locally on the onboard CPU and estimate the robot's pose from an AprilTag mounted on the robot, which is tracked by an external camera placed roughly eight meters above the ground (see \autoref{fig:real_world}(a)). Using this pose estimate and the wall positions pre-measured in a fixed maze frame, we sample the height map around the robot at every time step, reproducing the height-scan observation from simulation without onboard terrain perception.
The policy does not receive a goal or any an externally generated navigation command and explores the maze fully autonomously.
It selects directions at intersections, traverses longer corridors, and continues exploring after returning to previously visited parts of the maze.
The snapshot of the matrix memory in \autoref{fig:real_world} shows how the policy modifies the cells along its trajectory while cells outside the visited region retain their initial value. The same structure is visible in the time spent in each cell shown in \autoref{fig:heat_map_real}, where the regions that received no matrix memory updates are also
the ones the robot never entered. However, the visit distribution covers enough of the maze to reduce the reconstruction error similarly to the simulation results, which is shown in \autoref{fig:bce_real}. As the maze is significantly smaller than in the simulation setup, the policy observes a higher fraction of it early on, which leads to a faster error reduction. During deployment runs, we noticed that the policy sometimes tries to cut corners or even touch the walls, which we could only reduce with strong penalties on cell centering and wall proximity.
This is necessary in our more fragile real world maze setup, but with proper solid walls it might be beneficial to let the policy touch its surroundings as another form of re-orientation and grounding of its predictions, as long as the impact forces are mild.

\section{Limitations \& Future work}
The \gls{lsmm} memory covers a fixed area that is aligned with the environment
frame and sized to the maze, so exploring environments of unknown sizes would
require to grow or shift the memory with the robot.
The update further depends on an external position estimate, and although
training with $\pm0.20\,\mathrm{m}$ of noise makes the policy tolerate
imperfect localization, we do not yet estimate the position from onboard
sensing alone.
Finally, our mazes are flat and small enough that reactive strategies
reach similar coverage, so we expect the benefit of a spatial memory for
pure exploration to grow with the size of the environment.
In future work, we want to scale \gls{lsmm} to larger and rougher
environments, estimate the position internally from the robot's own
sensing, and use the reconstructed map for downstream tasks such as
navigating back to previously seen locations.

\section{Conclusion}

We introduced \gls{lsmm}, a spatially indexed latent memory that lets a
single policy explore unseen mazes end-to-end, from onboard observations
to joint-level actions.
By applying the \gls{hmf} update only to a local window centered on the
robot's position estimate, information about a location persists in the
memory until the robot returns there.
In our experiments, unstructured recurrent and finite-history baselines
match \gls{lsmm} in pure maze coverage, but they explore reactively and
forget the layout they have seen.
\gls{lsmm} instead builds a persistent map of the maze while exploring,
reduces the reconstruction error by a factor of three, spreads its visits most
evenly over the maze, and additionally stores physical properties of the
environment.
We deployed the policy on a Unitree Go2 and demonstrated goal-free
exploration of a real outdoor maze without an explicit map, planner, or
waypoint commands.

\section*{APPENDIX}

To ensure a fair comparison, we train all memory architectures in our
experiments with the same learning hyperparameters, which are listed in
\autoref{tab:hyperparameters}.

\begin{table}[h]
\setlength{\tabcolsep}{5pt}
\def\arraystretch{1.1}
\caption{Hyperparameters used in all experiments.}
\label{tab:hyperparameters}
\centering
\footnotesize
\begin{tabular}{@{}lr@{}}
\toprule
\textbf{Hyperparameter} & \textbf{Value} \\
\midrule
Training steps & $2.5 \cdot 10^{9}$ \\
Batch size (envs $\times$ steps per env) & $5\,000 \times 20 = 100\,000$ \\
Minibatch size / updates per batch & $25\,000$ / $5$ \\
\midrule
Discount factor $\gamma$ / GAE $\lambda$ & $0.995$ / $0.9$ \\
Clipping range & $0.2$ \\
Entropy coef.\ (linearly annealed) & $4 \cdot 10^{-4} \rightarrow 10^{-4}$ \\
\midrule
Actor optimizer, learning rate & Adam, $3 \cdot 10^{-4}$ \\
Critic optimizer, learning rate & AdamW, $5 \cdot 10^{-4}$ \\
Vel.\ estimator \& wall predictor opt., lr & AdamW, $3 \cdot 10^{-4}$ \\
Gradient clip norm & $1.0$ \\
Weight decay (AdamW) & $10^{-4}$ \\
\midrule
Loss weights $\lambda_{\text{maze}}$ / $\lambda_{\text{vel}}$ & $0.5$ / $0.1$ \\
\bottomrule
\end{tabular}
\end{table}
\autoref{tab:rewards} lists all reward terms and their coefficients.
All terms are scaled with their coefficients and summed up, and the result is
clipped to be above zero before the constant alive bonus is added
outside the clipping.
\begin{table}[t]
\setlength{\tabcolsep}{5pt}
\def\arraystretch{1.1}
\caption{Reward terms and coefficients, with $[x]_+ = \max(x,0)$ and
$\overline{\,\cdot\,}$ the mean over joints.
E/G: exploration/goal episodes only;
$n_{\text{cell}}$: steps in the current cell;
$\delta_i$: offset from the cell center toward walled side $i$;
$n^{\text{air}}_{\text{pairs}}$: leg pairs with both feet airborne;
$t^{\text{air}}$: foot air time.}
\label{tab:rewards}
\centering
\footnotesize
\begin{tabular}{@{}llr@{}}
\toprule
\textbf{Reward} & \textbf{Term} & \textbf{Coeff.} \\
\midrule
\multicolumn{3}{@{}l}{\textit{Task}} \\
Cell visit (E) & $\exp(-n_{\text{cell}}/200)$ & $0.16$ \\
Maze coverage (E) & fraction of visited cells & $0.1$ \\
Goal direction (G) & $\tfrac{1}{2}\big(1{+}\cos(\mathbf{v}_{xy},\mathbf{g}_t)\big)
\min(\lVert\mathbf{v}_{xy}\rVert,1)$ & $0.04$ \\
Goal reached (G) & $\mathds{1}[d_{\text{goal}}{<}0.25]$ & $0.0625$ \\
\midrule
\multicolumn{3}{@{}l}{\textit{Navigation shaping}} \\
Alive & $1$ & $0.03$ \\
Stalling & $-[0.1-\lVert\mathbf{v}_{xy}\rVert]_+$ & $0.1$ \\
Wall proximity & $-\mathds{1}[d_{\text{wall}}{<}0.35]\,
e^{-5(d_{\text{wall}}-0.15)}$ & $0.015$ \\
Cell centering & $-\sum_i [\delta_i]_+^2$ & $0.04$ \\
\midrule
\multicolumn{3}{@{}l}{\textit{Gait shaping}} \\
Vertical velocity & $-v_z^2$ & $0.04$ \\
Roll-pitch velocity & $-\lVert\boldsymbol{\omega}_{xy}\rVert^2$ & $10^{-3}$ \\
Roll-pitch position & $-\lVert\boldsymbol{\phi}_{xy}\rVert^2$ & $0.2$ \\
Joint position limits & $-\overline{[\mathbf{q}-\mathbf{q}^{\max}]_+
+ [\mathbf{q}^{\min}-\mathbf{q}]_+}$ & $0.8$ \\
Joint velocity limits & $-\overline{[\lvert\dot{\mathbf{q}}\rvert
- 0.9\,\dot{\mathbf{q}}^{\max}]_+}$ & $0.1$ \\
Joint velocity & $-\overline{\dot{\mathbf{q}}^2}$ & $8 \cdot 10^{-6}$ \\
Joint acceleration & $-\overline{\ddot{\mathbf{q}}^2}$ & $10^{-7}$ \\
Joint torque & $-\overline{(\boldsymbol{\tau}/\boldsymbol{\tau}^{\max})^2}$
& $6 \cdot 10^{-4}$ \\
Power draw & $-\overline{[\boldsymbol{\tau}\dot{\mathbf{q}}]_+
/(\boldsymbol{\tau}^{\max}\dot{\mathbf{q}}^{\max})}$ & $6 \cdot 10^{-4}$ \\
Action rate & $-\overline{(\mathbf{a}_t-\mathbf{a}_{t-1})^2}$ & $0.06$ \\
Nominal position & $-\overline{(\mathbf{q}-\mathbf{q}^{\text{nom}})^2}$ & $0.2$ \\
Base height & $-\min\big((h-h^{\text{nom}})^2,1\big)$ & $0.6$ \\
Air time & $-\overline{[0.3-t^{\text{air}}]_+}$ & $0.06$ \\
Feet pair air & $-n^{\text{air}}_{\text{pairs}}$ & $0.01$ \\
Ground collision & $-\mathds{1}[\text{body below ground}]$ & $2.0$ \\
\bottomrule
\end{tabular}
\end{table}

\addtolength{\textheight}{-6cm}
\IEEEtriggeratref{20} 
\bibliographystyle{IEEEtran}
\bibliography{bibliography}

@article{schulman2017proximal,
  title={Proximal policy optimization algorithms},
  author={Schulman, John and Wolski, Filip and Dhariwal, Prafulla and Radford, Alec and Klimov, Oleg},
  journal={arXiv preprint arXiv:1707.06347},
  year={2017}
}

@inproceedings{le2025stable,
  title={Stable hadamard memory: Revitalizing memory-augmented agents for reinforcement learning},
  author={Le, Hung and Nguyen, Dung and Do, Kien and Gupta, Sunil and Venkatesh, Svetha},
  booktitle={International Conference on Learning Representations},
  volume={2025},
  pages={73455--73472},
  year={2025}
}

@inproceedings{rudin2022advanced,
  title={Advanced skills by learning locomotion and local navigation end-to-end},
  author={Rudin, Nikita and Hoeller, David and Bjelonic, Marko and Hutter, Marco},
  booktitle={2022 IEEE/RSJ International Conference on Intelligent Robots and Systems (IROS)},
  pages={2497--2503},
  year={2022},
  organization={IEEE}
}

@inproceedings{zhuang2023,
  author =       {Zhuang, Ziwen and Fu, Zipeng and Wang, Jianren and Atkeson,
                  Christopher and Schwertfeger, Sören and Finn, Chelsea and
                  Zhao, Hang},
  title =        {Robot Parkour Learning},
  booktitle =    {Conference on Robot Learning ({CoRL})},
  year =         2023,
}

@inproceedings{cheng2023,
  title =        {Extreme Parkour with Legged Robots},
  author =       {Cheng, Xuxin and Shi, Kexin and Agarwal, Ananye and Pathak,
                  Deepak},
  booktitle =    {RoboLetics: Workshop on Robot Learning in Athletics@ CoRL
                  2023},
  year =         2023
}

@article{hoeller2024anymal,
  title={Anymal parkour: Learning agile navigation for quadrupedal robots},
  author={Hoeller, David and Rudin, Nikita and Sako, Dhionis and Hutter, Marco},
  journal={Science Robotics},
  volume={9},
  number={88},
  pages={eadi7566},
  year={2024},
  publisher={American Association for the Advancement of Science}
}

@article{ji2022concurrent,
  title={Concurrent training of a control policy and a state estimator for dynamic and robust legged locomotion},
  author={Ji, Gwanghyeon and Mun, Juhyeok and Kim, Hyeongjun and Hwangbo, Jemin},
  journal={IEEE Robotics and Automation Letters},
  volume={7},
  number={2},
  pages={4630--4637},
  year={2022},
  publisher={IEEE}
}

@article{hochreiter1997long,
  title={Long short-term memory},
  author={Hochreiter, Sepp and Schmidhuber, J{\"u}rgen},
  journal={Neural computation},
  volume={9},
  number={8},
  pages={1735--1780},
  year={1997},
  publisher={MIT press}
}

@inproceedings{cho2014properties,
  title={On the properties of neural machine translation: Encoder--decoder approaches},
  author={Cho, Kyunghyun and Van Merri{\"e}nboer, Bart and Bahdanau, Dzmitry and Bengio, Yoshua},
  booktitle={Proceedings of SSST-8, eighth workshop on syntax, semantics and structure in statistical translation},
  pages={103--111},
  year={2014}
}

@article{vaswani2017attention,
  title={Attention is all you need},
  author={Vaswani, Ashish and Shazeer, Noam and Parmar, Niki and Uszkoreit, Jakob and Jones, Llion and Gomez, Aidan N and Kaiser, {\L}ukasz and Polosukhin, Illia},
  journal={Advances in neural information processing systems},
  volume={30},
  year={2017}
}

@inproceedings{yamauchi1997frontier,
  title={A frontier-based approach for autonomous exploration},
  author={Yamauchi, Brian},
  booktitle={Proceedings 1997 IEEE International Symposium on Computational Intelligence in Robotics and Automation CIRA'97.'Towards New Computational Principles for Robotics and Automation'},
  pages={146--151},
  year={1997},
  organization={IEEE}
}

@article{mirowski2016learning,
  title={Learning to navigate in complex environments},
  author={Mirowski, Piotr and Pascanu, Razvan and Viola, Fabio and Soyer, Hubert and Ballard, Andrew J and Banino, Andrea and Denil, Misha and Goroshin, Ross and Sifre, Laurent and Kavukcuoglu, Koray and others},
  journal={arXiv preprint arXiv:1611.03673},
  year={2016}
}

@inproceedings{zhang2024resilient,
  title={Resilient legged local navigation: Learning to traverse with compromised perception end-to-end},
  author={Zhang, Chong and Jin, Jin and Frey, Jonas and Rudin, Nikita and Mattamala, Mat{\'\i}as and Cadena, Cesar and Hutter, Marco},
  booktitle={2024 IEEE International Conference on Robotics and Automation (ICRA)},
  pages={34--41},
  year={2024},
  organization={IEEE}
}

@article{mattamala2025wild,
  title={Wild visual navigation: Fast traversability learning via pre-trained models and online self-supervision},
  author={Mattamala, Matias and Frey, Jonas and Libera, Piotr and Chebrolu, Nived and Martius, Georg and Cadena, Cesar and Hutter, Marco and Fallon, Maurice},
  journal={Autonomous Robots},
  volume={49},
  number={3},
  pages={19},
  year={2025},
  publisher={Springer}
}

@inproceedings{miki2022elevation,
  title={Elevation mapping for locomotion and navigation using gpu},
  author={Miki, Takahiro and Wellhausen, Lorenz and Grandia, Ruben and Jenelten, Fabian and Homberger, Timon and Hutter, Marco},
  booktitle={2022 IEEE/RSJ International Conference on Intelligent Robots and Systems (IROS)},
  pages={2273--2280},
  year={2022},
  organization={IEEE}
}

@article{bohlinger2025gait,
  title={Gait in Eight: Efficient On-Robot Learning for Omnidirectional Quadruped Locomotion},
  author={Bohlinger, Nico and Kinzel, Jonathan and Palenicek, Daniel and Antczak, Lukasz and Peters, Jan},
  journal={arXiv preprint arXiv:2503.08375},
  year={2025}
}

@inproceedings{aditya2025robust,
  title={Robust localization, mapping, and navigation for quadruped robots},
  author={Aditya, Dyuman and Huang, Junning and Bohlinger, Nico and Kicki, Piotr and Walas, Krzysztof and Peters, Jan and Luperto, Matteo and Tateo, Davide},
  booktitle={2025 European Conference on Mobile Robots (ECMR)},
  pages={1--8},
  year={2025},
  organization={IEEE}
}

@inproceedings{stasica2025bridge,
  title={Bridge the gap: Enhancing quadruped locomotion with vertical ground perturbations},
  author={Stasica, Maximilian and Bick, Arne and Bohlinger, Nico and Mohseni, Omid and Fritzsche, Max Johannes Alois and H{\"u}bler, Clemens and Peters, Jan and Seyfarth, Andr{\'e}},
  booktitle={2025 IEEE/RSJ International Conference on Intelligent Robots and Systems (IROS)},
  pages={13370--13370},
  year={2025},
  organization={IEEE}
}

@article{bohlinger2024onepolicy,
    title={One Policy to Run Them All: an End-to-end Learning Approach to Multi-Embodiment Locomotion},
    author={Bohlinger, Nico and Czechmanowski, Grzegorz and Krupka, Maciej and Kicki, Piotr and Walas, Krzysztof and Peters, Jan and Tateo, Davide},
    journal={Conference on Robot Learning},
    year={2024}
}

@article{lecun1989backpropagation,
  title={Backpropagation applied to handwritten zip code recognition},
  author={LeCun, Yann and Boser, Bernhard and Denker, John S and Henderson, Donnie and Howard, Richard E and Hubbard, Wayne and Jackel, Lawrence D},
  journal={Neural computation},
  volume={1},
  number={4},
  pages={541--551},
  year={1989},
  publisher={MIT Press}
}

@inproceedings{parisotto2018neural,
  title={Neural Map: Structured Memory for Deep Reinforcement Learning},
  author={Parisotto, Emilio and Salakhutdinov, Ruslan},
  booktitle={International Conference on Learning Representations},
  year={2018}
}

@inproceedings{henriques2018mapnet,
  title={Mapnet: An allocentric spatial memory for mapping environments},
  author={Henriques, Joao F and Vedaldi, Andrea},
  booktitle={proceedings of the IEEE Conference on Computer Vision and Pattern Recognition},
  pages={8476--8484},
  year={2018}
}

@article{chaplot2020learning,
  title={Learning to explore using active neural slam},
  author={Chaplot, Devendra Singh and Gandhi, Dhiraj and Gupta, Saurabh and Gupta, Abhinav and Salakhutdinov, Ruslan},
  journal={arXiv preprint arXiv:2004.05155},
  year={2020}
}

@inproceedings{al2024time,
  title={Time-efficient reinforcement learning with stochastic stateful policies},
  author={Al-Hafez, Firas and Zhao, Guoping and Peters, Jan and Tateo, Davide},
  booktitle={International Conference on Learning Representations},
  volume={2024},
  pages={49532--49568},
  year={2024}
}

@inproceedings{liulocoformer,
  title={LocoFormer: Generalist Locomotion via Long-context Adaptation},
  author={Liu, Min and Pathak, Deepak and Agarwal, Ananye},
  booktitle={9th Annual Conference on Robot Learning},
  year={2025}
}

@inproceedings{pinto2018asymmetric,
  title={Asymmetric actor critic for image-based robot learning},
  author={Pinto, Lerrel and Andrychowicz, Marcin and Welinder, Peter and Zaremba, Wojciech and Abbeel, Pieter},
  booktitle={Proceedings of Robotics: Science and Systems (RSS)},
  year={2018},
}

@inproceedings{todorov2012mujoco,
  title={MuJoCo: A physics engine for model-based control},
  author={Todorov, Emanuel and Erez, Tom and Tassa, Yuval},
  booktitle={2012 IEEE/RSJ International Conference on Intelligent Robots and Systems},
  pages={5026--5033},
  year={2012},
  organization={IEEE},
  doi={10.1109/IROS.2012.6386109}
}

@article{ai2025towards,
  title={Towards Embodiment Scaling Laws in Robot Locomotion},
  author={Ai, Bo and Dai, Liu and Bohlinger, Nico and Li, Dichen and Mu, Tongzhou and Wu, Zhanxin and Fay, K and Christensen, Henrik I and Peters, Jan and Su, Hao},
  journal={Conference on Robot Learning (CoRL)},
  year={2025}
}

@article{bohlinger2025multi,
  title={Multi-Embodiment Locomotion at Scale with extreme Embodiment Randomization},
  author={Bohlinger, Nico and Peters, Jan},
  journal={arXiv preprint arXiv:2509.02815},
  year={2025}
}

@article{li2026online,
  title={Rapid Embodiment Adaptation for Quadrupedal Locomotion},
  author={Li, Dichen and Ai, Bo and Bohlinger, Nico and Peters, Jan and Su, Hao and Christensen, Henrik I},
  journal={arXiv preprint arXiv:2608.01506},
  year={2026}
}

@article{yang2025spatially,
  title={Spatially-Enhanced Recurrent Memory for Long-Range Mapless Navigation via End-to-End Reinforcement Learning},
  author={Yang, Fan and Frivik, Per and Hoeller, David and Wang, Chen and Cadena, Cesar and Hutter, Marco},
  journal={International Journal of Robotics Research},
  year={2025},
  archivePrefix={arXiv},
  eprint={2506.05997},
  primaryClass={cs.RO}
}

\end{document}